%% file: acl_latex.tex
\pdfoutput=1
\documentclass[11pt]{article}

\usepackage[preprint]{acl}

\usepackage{times}
\usepackage{latexsym}

\usepackage[T1]{fontenc}
\usepackage[utf8]{inputenc}

\usepackage{microtype}

\usepackage{inconsolata}

\usepackage{graphicx}

\usepackage{url}
\usepackage{multirow}
\usepackage{csquotes}
\usepackage{siunitx}
\usepackage{fontawesome}
\usepackage{tikz}

\title{Detection of fields of applications in biomedical abstracts \\ with the support of argumentation elements}

\author{Mariana Neves \\
  German Centre for the Protection of Laboratory Animals (Bf3R), \\
  German Federal Institute for Risk Assessment (BfR), Berlin, Germany\\
  \texttt{mariana.lara-neves@bfr.bund.de} \\}

\begin{document}
\maketitle
\begin{abstract}
\input{0_Abstract}

\end{abstract}

\section{Introduction}
\label{sec:introduction}
\input{1_Introduction.tex}

\section{Tools for Argument Mining}
\label{sec:argumentmining}

\input{2_Argument_Mining_Tools}

\section{Corpus and Text Classification}
\label{sec:datamethods}

\input{3_DataMethods}

\section{Results and Discussion}
\label{sec:results}

\input{4_Results}

\section{Conclusions}
\input{6_Conclusions}

\bibliography{references}

\end{document}

%% file: 0_Abstract.tex
Focusing on particular facts, instead of the complete text, can potentially improve searching for specific information in the scientific literature. 
In particular, argumentative elements allow focusing on specific parts of a publication, e.g., the background section or the claims from the authors.
We evaluated some tools for the extraction of argumentation elements for a specific task in biomedicine, namely, for detecting the fields of the application in a biomedical publication, e.g, whether it addresses the problem of disease diagnosis or drug development.
We performed experiments with the PubMedBERT pre-trained model, which was fine-tuned on a specific corpus for the task.
We compared the use of title and abstract to restricting to only some argumentative elements.
The top F1 scores ranged from 0.22 to 0.84, depending on the field of application.
The best argumentative labels were the ones related the conclusion and background sections of an abstract.

%% file: 1_Introduction.tex
Searching for specific information in the biomedical literature is a time consuming and complex task, for which text mining approaches have been developed in the recent years \citep{KRALLINGER2005439}.
Current tools allow to enrich the articles with automatically predicted information with an adequate quality, e.g., for a variety of entity types (e.g., PubTator \citealp{10.1093/nar/gkz389}) and relations \citep{10.1093/bib/bbac282}, or for classifying the text according to pre-defined labels, e.g., for chemical exposure \citep{10.1371/journal.pone.0173132} or hallmarks of cancer \citep{10.1093/bioinformatics/btv585}.

However, there are few studies on the extraction of argumentative elements from biomedical publications \citep{10.1002/asi.24590}.
While argumentative elements might not be relevant for the extraction of many biomedical facts, e.g., gene/protein names, they can potentially support mining facts that appear in some particular parts of a publication, which is the case of detecting the fields of application in a publication.
For instance, articles about breast cancer may refer to different fields of applications, e.g., understanding the disease mechanism \citep{10.12688/f1000research.15447.2}, or contributing to its diagnosis \citep{KATTI2017125} or treatment \citep{Pradhan2018}.

However, the detection of the application is a hard task, since it requires a detailed understanding of the text (cf. Table~\ref{tab:examples}), and sometimes even the use of the full text.
For instance, in an article about the diagnosis of breast cancer \citep{KATTI2017125}, the authors describe that osteoblasts and other bone cells can support the diagnosis, since breast cancer cells can migrate to the bones.
Ideally, the extraction of such complex facts should rely on a combination of various named entities (e.g., \citealp{10.1093/nar/gkz389}) and relations, which are not always readily available.
However, since the relevant information is mentioned in a couple of sentences of the abstract, we hypothesise that it can be supported by specific argumentative elements.

We aimed at evaluating particular argumentative elements, which we extracted using available tools.
We addressed the detection of fields of applications as a text classification task and relied on the titles, abstracts and the argumentative elements.
We trained and evaluated it on a new corpus of more than 2k abstracts that contains eight labels for the fields of application.
The corpus was compiled from a series of seven reports about advanced non-animal models in biomedical research, e.g., \citep{doi/10.2760/725821}.

%We are not aware of any previous similar corpus.
%nor many previous publications that address the topic.
%While many previous research focused on the extraction of various entity types (??), 
%360	knowledge_base
%197	target_discovery
%87	clinical_finding
%25	biological_function
%38	drug_discovery
%18	method_development
%16	review
%2	preclinical_testing
%56	model_development
We are aware of only one previous work in which authors the annotated 387 abstracts with eight stages in biomedical research \citep{Butzke2020SMAFIRAcAB}, e.g.,  drug\_discovery, method\_development.
%knowledge\_base, target\_discovery and clinical\_finding, biological\_function, drug\_discovery, method\_development, preclinical\_testing, and model\_development.
Some of these categories overlap with the ones contained in the reports, %such as drug\_discovery or method\_development.
however, their data is too short for training purposes.

Our contributions are the following:
(i) a new corpus (which will be released) of around 2k biomedical abstracts and eight labels for the fields of applications;
(ii) a short review of the available tools for the extraction of argumentative elements;
(iii) an evaluation of these tools for the task of detecting fields of applications in biomedical abstracts;
(iv) and the annotations for our corpus from various tools for argumentation.

%% file: 2_Argument_Mining_Tools.tex
There are many previous studies on argument mining for scientific articles \citep{10.1002/asi.24590}.
However, for the field of biomedicine, few studies have been carried out and few corpora or tools have been made available.
% https://repositori.upf.edu/handle/10230/47600
% https://ceur-ws.org/Vol-2847/paper-03.pdf
% https://github.com/LaSTUS-TALN-UPF/SciARG
\citet{AccuostoEtAl:BIR2021} carried out the annotation of the SciARG corpus, which includes various units (e.g., \enquote{proposal}, \enquote{observation}) and relations (e.g., \enquote{elaborate} or \enquote{support}).
The corpus contains articles for computational linguistics and biomedicine, but as far as we know, no tool is available for use.
Some corpora built for semantic annotation in biomedicine also include annotations that are relevant for the field of argumentation mining.
% https://www.frontiersin.org/articles/10.3389/frma.2021.674205/full
The ECO-CollecTF corpus \citep{10.3389/frma.2021.674205} contains the assertion strength (high, medium, or low), as expressed in the text (e.g., words \enquote{conclude} or \enquote{possibly}).
% ACTA: Argumentative Clinical Trial Analysis
Further, ACTA \citep{ijcai2019p953} was developed for mining arguments for clinical trials, and was used for exploring the COVID-19 scientific literature \citep{10.1162/qss_a_00164}.
However, we only found an online demo version that is not suitable for batch processing.
% https://direct.mit.edu/qss/article/2/4/1301/108049/Covid-on-the-Web-Exploring-the-COVID-19-scientific
% https://www.ijcai.org/proceedings/2019/953

We searched the scientific literature for tools for the extraction of argumentative elements.
We did not restrict them for the biomedical domain nor on their suitability for scientific articles.
We included tools for discourse elements, even though these do not really constitute argumentation elements.
However, as discussed in \citep{LIPPI2016292}, very few tools are readily available for argumentation mining.
We provide an overview of the tools that we found and tried.

% http://ltdemos.informatik.uni-hamburg.de/targer/
% http://aclanthology.lst.uni-saarland.de/P19-3031.pdf
We ran TARGER \citep{chernodub-etal-2019-targer} on our corpus, but did not consider its annotations in our experiments.
% (cf. Section~\ref{sec:datamethods})
For our corpus, practically all abstracts were tagged as \enquote{premise}, with only one passage\footnote{From PMID 29133340: \enquote{Cardiac arrhythmias are major life-threatening conditions}} tagged as \enquote{claim}.
% https://www.zora.uzh.ch/id/eprint/217835/1/2022_SDU_AAAI_RuoschEtAl__CEUR_.pdf
% https://gitlab.ifi.uzh.ch/DDIS-Public/bam
We installed BAM \citep{zora217835}, but the tool first needs to be trained in order to be used for tagging, after previously downloading the respective corpora.
% https://ojs.aaai.org/index.php/AAAI/article/download/6438/6294
% https://github.com/trtm/AURC
Similarly, after the installation, AURC \citep{epub88715} failed to download the necessary data for its training, and the code seems not be ready for tagging any given text.
% https://aclanthology.org/2021.argmining-1.10.pdf
% https://arxiv.org/pdf/2210.13084.pdf
% https://webis.de/downloads/publications/papers/alkhatib_2021b.pdf
% https://github.com/afergadis/SciARK
We also failed to make SciARK \citep{fergadis-etal-2021-argumentation} work due to various problems with the libraries.
Further, we missed some clear instructions on how to train the model in order to get the best performing one.
%args.me
% https://www.args.me/index.html
% https://aclanthology.org/W17-5106/
%ArgumenText
% https://www.argumentsearch.com/
% https://aclanthology.org/N18-5005/
Finally, we excluded arg.me \citep{wachsmuth-etal-2017-building} and ArgumenText \citep{stab-etal-2018-argumentext}, since they do not allow to process any given text.
In summary, only three tools worked succesfully and could be used in our experiments, as listed below:

% https://aclanthology.org/W18-5203.pdf
% https://github.com/anlausch/ArguminSci
\paragraph{ArguminSci.}
It was trained on the Dr. Inventor corpus and includes five schemes \citep{lauscher-etal-2018-arguminsci}, from which we selected the following ones: (i) argumentative components, i.e., \enquote{Background claim} and \enquote{Own claim}; and (ii) discourse roles, i.e., \enquote{Background}, \enquote{Challenge}, \enquote{Approach}, \enquote{Outcome}, and \enquote{Future work}.
%(3) subjective aspect classification, (4) summary relevance classification, and (5) citation context identification.
The tool is available for download\footnote{\url{https://github.com/anlausch/ArguminSci}}, as well as the models.

% https://aclanthology.org/D18-1349/
\paragraph{HSLN.}
It provides discourse elements and was trained on a large collection of biomedical abstracts from PubMed \citep{jin-szolovits-2018-hierarchical}.
Its source code is freely available for download\footnote{\url{https://github.com/jind11/HSLN-Joint-Sentence-Classification}} and it provides annotations for the following elements: \enquote{Background}, \enquote{Objective}, \enquote{Methods}, \enquote{Results}, and \enquote{Conclusion}.

% Margot
% https://www.sciencedirect.com/science/article/abs/pii/S0957417416304493
% http://margot.disi.unibo.it/ESWA2016.pdf
\paragraph{MARGOT.}
It was trained on Wikipedia articles and returns two types of annotations: \enquote{Claim} and \enquote{Evidence}.
The tool is online available\footnote{\url{margot.disi.unibo.it/}} and as a Web service in the Penelope suite\footnote{\url{https://app.swaggerhub.com/apis/EHAI/penelope-margot-service/1.0.0}}.
%, and for download (under request).
% https://www.ncbi.nlm.nih.gov/pmc/articles/PMC9339778/
% https://www.frontiersin.org/articles/10.3389/fpubh.2022.945181/full
% https://www.ijcai.org/proceedings/2022/0857.pdf
It was previously used for mining claims and evidences for publications about COVID-19 \citep{10.3389/fpubh.2022.945181,ijcai2022p857}.

%% file: 3_DataMethods.tex
We derived a corpus from seven reports by the European Commission about advanced non-animal models in biomedical research.
The reports provide a summary of available non-animal models that can potentially be used in biomedical research.
Seven reports were published for seven topics, namely, 
respiratory tract diseases (RTD) \citep{doi/10.2760/725821},
breast cancer (BC) \citep{doi/10.2760/618741},
neurodegenerative diseases (ND) \citep{doi/10.2760/386},
immuno-oncology (IO) \citep{doi/10.2760/393670},
immunogenicity testing for advanced therapy medicinal products (IMM) \citep{10.2760/7190},
cardiovascular diseases (CD) \citep{doi/10.2760/94608} and
autoimmune diseases (AID) \citep{doi/10.2760/617688}.
Table~\ref{tab:examples} shows examples for some applications and Table~\ref{tab:app_reports} presents the number of applications per report.

\begin{table*}[t!]
\small
\begin{center}
\begin{tabular}{|c|p{14.5cm}|}
\hline
{\bf Exam.} & {\bf Passages} \\
\hline
1 & \enquote{Overexpression of S6K1 was detected in [...] breast cancer, and correlated with the worse disease outcome. [...] suggesting the implication of kinase nuclear substrates in tumor progression.} \\
\hline
2 & \enquote{softer substrates such as osteoblasts and other bone cells result in a much altered unloading response as well as significant plastic deformation. These substrates are relevant to metastasis...} \\
\hline
3 & \enquote{Assessment of anti-cancer drug efficacy in in vitro three-dimensional (3D) bioengineered cancer models provides [...] towards pre-clinical translation of potential drug candidates.} \\
\hline
\end{tabular}
\end{center}
\caption{Examples with passages that support the applications: (1)  \enquote{disease\_mechanism\_exp\_theor} (PMID 30705751), (2) \enquote{diagnosis\_of\_disease} (PMID 28571747), and (3) \enquote{disease\_therapy\_develop} (PMID 29453454).
}
\label{tab:examples}
\end{table*}

Each report contains a corresponding spreadsheet in which the reviewed articles are listed along with details about the methods, field of application, disease, biological endpoints, among others.
From each spreadsheet, we collected the article title, DOI, and the corresponding field(s) of application.
We used the DOIs to query for the articles in PubMed (if available) using the Entrez Programming Utilities\footnote{\url{https://www.ncbi.nlm.nih.gov/books/NBK25497/}}, and obtained a total of 2,859 citations (PMIDs).

%\paragraph{Field of applications.}
Most reports contain roughly the same seven fields of application, even though sometimes they appeared with slightly different names across the reports.
However, two of the reports, namely \enquote{respiratory tract diseases} and \enquote{neurodegenerative diseases}, include some topic-specific applications, e.g., \enquote{mimics lung microenvironment}.
We unified these disease-specific applications under the application \enquote{Others}.
The column \enquote{Corpus} in Table~\ref{tab:app_reports} summarizes the frequency of each application in the derived corpus.

% all_apps
%1217	Disease_mechanism_exp_theor
%458	Drug_develop_testing
%642	Disease_therapy_develop
%296	Model_method_develop_exp
%61	Others
%75	Model_method_qualification
%95	Model_method_develop_theor
%199	Diagnosis_of_disease
\begin{table*}[t!]
\small
\begin{center}
\begin{tabular}{|l|c|c|c|c|c|c|c|c|c|}
\hline
{\bf Applications/Labels} & {\bf Abbrev.} & {\bf RTD} & {\bf BC} & {\bf ND} & {\bf IO} & {\bf IMM} & {\bf CD} & {\bf AID} & {\bf Corpus} \\
\hline
%Diagnosis_of_disease	1:3	2:21	3:87	4:6	6:97	7:4
Diagnosis\_of\_disease & DgDs & 3 & 21 & 87 & 6 & 0 & 97 & 4 & 199 \\
%Disease_mechanism_exp_theor	1:79	2:475	3:287	4:214	6:127	7:71
Disease\_mechanism\_exp\_theor & DsMc & 79 & 475 & 287 & 214 & 0 & 127 & 71 & 1217 \\
%Disease_therapy_develop	1:28	2:22	3:67	4:205	5:79	6:222	7:32
Disease\_therapy\_develop & DsTr & 28 & 22 & 67 & 205 & 79 & 222 & 32 & 642 \\
%Drug_develop_testing	1:32	2:285	3:24	4:88	7:42
Drug\_develop\_testing & DrDv & 32 & 285 & 24 & 88 & 0 & 0 &	42 & 458 \\
%Model_method_develop_exp	1:89	2:154	3:51	4:14	5:2
Model\_method\_develop\_exp & MdEx & 89 & 154 & 51 & 14 & 2 & 0 & 0 & 296 \\
%Model_method_develop_theor	1:9	2:30	3:21	4:44	5:2
Model\_method\_develop\_theor & MdTh & 9 & 30 & 21 & 44 & 2 & 0 & 0 & 95 \\
%Model_method_qualification	2:36	4:2	5:5	7:32
Model\_method\_qualification & MdQl & 0 & 36 & 0 & 2 & 5 & 0 & 32 & 75 \\
%Others	1:42	3:23
Others & Othr & 42 & 0 & 23 & 0 & 0 & 0 & 0 & 61 \\
\hline
\end{tabular}
\end{center}
\caption{Number of articles for each application in the reports. 
%The reports are the following: respiratory tract diseases (RTD), breast cancer (BC), neurodegenerative diseases (ND), immuno-oncology (IO), immunogenicity testing for advanced therapy medicinal products (IMM), cardiovascular diseases (CD), and autoimmune diseases (AID).
The last column corresponds to the size of the corpus, and it is not the sum of the respective row due to the overlap of articles across the reports.}
\label{tab:app_reports}
\end{table*}

We addressed the problem as a multiclass, multilabel text classification, since some articles contain more than one label (application).
We carried out the usual current procedure for most NLP tasks, i.e., we relied on a pre-trained model, namely, PubmedBERT \citep{pubmedbert}, and fine-tuned it with the corpus.
Most of the articles (2,707) contain only one label, 133 of them with 2 labels, 14 of them with 3 labels, two of them with 5 labels, and only one with 4 (PMID 25855820), 6 (PMID 34237259), or 8 (PMID 30311153) labels.
We tried various values for the hyperparameters:
(a) batch sizes of 32, 64, and 128;
(b) 5, 10, 20, and 30 epochs;
and (c) learning rate of 3e-05, 4e-05, 5e-05, 6e-05, and 7e-05.
We considered the title and the abstracts of the articles as our baseline experiment, upon which the best set of hyperparameters was decided.

%% file: 4_Results.tex
\input{4_5_tables}

For our experiments, we randomly split the corpus into training (80\%), development (10\%), and test (10\%) sets.
We ran experiments with the title and abstracts, as well as with each of the argumentative elements from the selected tools (cf. Section~\ref{sec:argumentmining}).
For all these experiments, we concatenated the passage(s) of the element into a single text, following their original order in the abstract.
When no annotation was available for a particular element, we considered the title and abstracts instead.

We present the results in Table~\ref{tab:tools-results}, 
ranked by the overall F1 score over all applications.
Our best results were obtained when relying on the following hyperparameters: batch size of 128, a learning rate of 6e-05, and 10 epochs. 

The corpus is very unbalanced with respect to the labels.
As expected, the ones with more training examples (cf. Table~\ref{tab:app_reports}) obtained a higher performance.
However, even though \enquote{Othr} contains less training examples than \enquote{MdQl} (61 vs. 75, respectively), it obtained higher results (F1 of 0.40 vs. 0.22, respectively, for HSLN conclusions).
Similarly, the applications with more data (DsMc) were not the one that obtained the highest F1 score.

Roughly half of the elements scored higher than the baseline (title and abstract).
In the last two columns of Table~\ref{tab:tools-results}, we show the number of articles and the length of the extracted text per element.
On the one hand, elements with few articles, e.g., \enquote{Arguminsci future work} and \enquote{MARGOT claim}, were in the middle of the table and obtained similar results to the baseline.
Indeed, they had to rely on the titles and the abstracts for many articles.
On the other hand, elements with many articles and long text, e.g., \enquote{Arguminsci background} and \enquote{HSLN results} not always obtained the best results.
Seemingly, some elements are more informative than other for detecting the applications.
For most of the labels, results were very poor when relying exclusively on the extracted elements (results now shown), i.e., when not relying on the title and abstract for elements without any annotation for a particular article.

We analyzed the overlap between the elements by comparing the offsets of the corresponding extracted text.
We depict this correlation in Table~\ref{tab:overlap}.
Curiously, not much correlation was found between the discourse elements from HSLN and ArguminSci, even though they are roughly equivalent. 
For instance, \enquote{ArguminSci background} obtained one of the highest F1 score, while \enquote{HSLN background} obtained one of the lowest ones (cf. Table~\ref{tab:tools-results}).
Interestingly, the \enquote{evidence} element from MARGOT highly correlated with all other elements, but obtained one of the lowest F1 scores.

We checked the errors for the top-scored element (ArguminSci conclusions).
From a total of 112 errors,  19 of them were partial errors, since one of the two expected labels was correctly predicted. 
In 23 of the errors, a similar label was predicted, e.g., \enquote{DsTr} (disease therapy development) instead of \enquote{DsMc} (disease mechanism).

%% file: 4_5_tables.tex
%\faCircleO
%\faDotCircleO
%\faCircle

\begin{table*}[t!]
\small
\begin{center}
\begin{tabular}{|l|l|c|c|c|c|c|c|c|c|c|c|c|}
\hline
{\bf Tools} & {\bf Elem.} & {\bf Overall} & {\bf DgDs} & {\bf DsMc} & {\bf DsTr} & {\bf DrDv} & {\bf MdEx} & {\bf MdTh} & {\bf MdQl} & {\bf Othr} & {\bf Doc.} & {\bf Len.} \\
\hline
HSLN & concl. & \bf 0.57 & 0.63 & \bf 0.73 & \bf 0.62 & \bf 0.84 & \bf 0.65 & 0.44 & \bf 0.22 & 0.40 & \faCircle & \faDotCircleO \\
ArgumSci & backg. & 0.51 & 0.47 & 0.64 & 0.47 & 0.74 & \bf 0.65 & 0.43 & 0.00 & \bf 0.67 & \faCircle & \faCircle \\
ArgumSci & bck. cl. & 0.51 & 0.62 & 0.63 & 0.60 & 0.74 & 0.57 & 0.40 & 0.00 & 0.50 & \faCircle & \faDotCircleO \\
ArgumSci & appr. & 0.51 & \bf 0.71 & 0.59 & 0.64 & 0.80 & 0.61 & 0.46 & \bf 0.22 & 0.00 & \faCircle & \faDotCircleO \\
ArgumSci & fut.w. & 0.50 & 0.50 & 0.64 & 0.48 & 0.78 & 0.62 & 0.33 & 0.00 & \bf 0.67 & \faCircleO & \faCircleO \\
HSLN & methd. & 0.49 & 0.56 & 0.69 & 0.55 & 0.79 & 0.56 & 0.25 & 0.00 & 0.50 & \faDotCircleO & \faDotCircleO \\
\hline
\multicolumn{2}{|l|}{title-abstract} & 0.48 & 0.67 & 0.71 & 0.56 & 0.81 & 0.56 & 0.55 & 0.00 & 0.00 & - & - \\
\hline
MARGOT & claim & 0.48 & 0.50 & 0.66 & 0.51 & 0.79 & 0.53 & 0.20 & 0.00 & \bf 0.67 & \faCircleO & \faCircleO \\
HSLN & objec. & 0.48 & 0.63 & 0.64 & \bf 0.62 & 0.81 & 0.53 & 0.40 & \bf 0.22 & 0.00 & \faCircleO & \faCircleO \\
ArgumSci & own cl. & 0.47 & 0.59 & 0.71 & 0.61 & \bf 0.84 & 0.55 & 0.29 & 0.20 & 0.00 & \faCircle & \faCircleO \\
ArgumSci & outc. & 0.47 & \bf 0.71 & 0.69 & 0.54 & 0.77 & 0.57 & 0.22 & \bf 0.22 & 0.00 & \faCircle & \faDotCircleO \\
HSLN & resul. & 0.46 & 0.59 & 0.67 & 0.48 & 0.80 & 0.56 & \bf 0.60 & 0.00 & 0.00 & \faCircle & \faCircle \\
HSLN & backg. & 0.43 & 0.63 & 0.69 & 0.55 & 0.83 & 0.55 & 0.2 & 0.00 & 0.00 & \faCircle & \faDotCircleO \\
ArgumSci & chall. & 0.42 & 0.52 & 0.68 & 0.61 & 0.69 & 0.53 & 0.36 & 0.00 & 0.00 & \faDotCircleO & \faCircleO \\ 
MARGOT & evid. & 0.40 & 0.50 & 0.65 & 0.57 & 0.74 & 0.53 & 0.20 & 0.00 & 0.00 & \faDotCircleO & \faDotCircleO \\
\hline
\end{tabular}
\end{center}
\caption{The results are in terms of F1 and ordered from the highest to lowest score (\enquote{Overall}).
The \enquote{Doc} column shows the number of articles that contain the label: \faCircleO~(low, \textless 1k), \faDotCircleO, \faCircle~(high, \textgreater 2k).
The \enquote{Len} column shows the average length of the extracted text for the label: \faCircleO~(short, \textless 200), \faDotCircleO, \faCircle~(long, \textgreater 500).
}
\label{tab:tools-results}
\end{table*}

\newcommand*{\pie}[1]{%
\begin{tikzpicture}
 \draw (0,0) circle (1ex);\fill (1ex,0) arc (0:#1:1ex) -- (0,0) -- cycle;
\end{tikzpicture}%
}

\begin{table*}[t!]
\small
\begin{center}
\begin{tabular}{|l|l|c|c|c|c|c|c|c|c|c|c|c|}
\hline
\multirow{2}{*}{\bf Tools}  & \multirow{2}{*}{\bf Elem.} & \multicolumn{5}{c|}{\bf ArguminSci} & \multicolumn{2}{c|}{\bf MARGOT} & \multicolumn{2}{c|}{\bf ArguminSci} \\
\cline{3-11}
&& chall. & appr. & outc. & fut.w. & backg. & evid. & claim & own cl. & backg. cl. \\
\hline
\multirow{5}{*}{\bf HSLN} 
& objec. & \pie{0} & \pie{90} & \pie{0} & \pie{0} & \pie{0} & \pie{0} & \pie{0} & \pie{0} & \pie{0} \\
& methd. & \pie{0} & \pie{90} & \pie{0} & \pie{0} &  \pie{0} & \pie{90} & \pie{0} & \pie{0} & \pie{0} \\
& resul. & \pie{0} & \pie{180} & \pie{90} & \pie{0} & \pie{0} & \pie{360} & \pie{0} & \pie{0} & \pie{90} \\
& concl. & \pie{0} & \pie{90} & \pie{0} & \pie{0} &  \pie{0} & \pie{90} & \pie{0} & \pie{0} & \pie{0} \\
& backg. & \pie{90} & \pie{270} & \pie{90} & \pie{0} & \pie{0} & \pie{180} & \pie{0} & \pie{90} & \pie{90} \\
\hline
\multirow{5}{*}{\bf ArguminSci} 
& chall. & - & - & - & - & - & \pie{0} & \pie{0} & \pie{0} & \pie{0} \\
& appr. & - & - & - & - & - & \pie{180} & \pie{0} & \pie{0} & \pie{0} \\
& outc. & - & - & - & - & - & \pie{180} & \pie{0} & \pie{0} & \pie{0} \\
& fut.w. & - & - & - & - & - & \pie{0} & \pie{0} & \pie{0} & \pie{0} \\
& backg. & - & - & - & - & - & \pie{360} & \pie{0} & \pie{90} & \pie{270} \\
\hline
\multirow{2}{*}{\bf MARGOT}
& evid. & - & - & - & - & - & - & - & \pie{90} & \pie{180} \\
& claim & - & - & - & - & - & - & - & \pie{0} & \pie{0} \\
%\hline
%own cl. \\
%bac. cl. \\
\hline
\end{tabular}
\end{center}
\caption{Degree of overlap between the elements in this order: \protect\pie{0} (low), \protect\pie{90}, \protect\pie{180}, \protect\pie{270}, \protect\pie{360} (high). We did not compute the overlap between labels from the same tool.}
\label{tab:overlap}
\end{table*}

%% file: 6_Conclusions.tex
We provided a summary of available tools for the extraction of argumentative elements and tried a selection of them for the task of classifying scientific abstracts with regard to the fields of application.
We hope that this short survey can help others when choosing argumentative tools for other similar tasks.

Further, we compiled a new corpus derived from available reports by the European Commission. 
As far as we know, there is no similar corpus (in size) for the fields of applications.
We expect that the corpus can also be of use for benchmarking text classification in the biomedical domain \citep{pubmedbert}, for which experiments are currently usually restricted to the corpus for hallmarks of cancer \citep{10.1093/bioinformatics/btv585}.
In our experiments, 
many of the elements outperformed the baseline and the predictions based on the \enquote{conclusion} element of the HSLN tool obtained the best results.

%% file: acl_latex.bbl
\begin{thebibliography}{33}
\providecommand{\natexlab}[1]{#1}

\bibitem[{Accuosto et~al.(2021)Accuosto, Neves, and
  Saggion}]{AccuostoEtAl:BIR2021}
Pablo Accuosto, Mariana Neves, and Horacio Saggion. 2021.
\newblock \href {http://ceur-ws.org/Vol-2847/#paper-03} {Argumentation mining
  in scientific literature: From computational linguistics to biomedicine}.
\newblock In \emph{Proceedings of the BIR 2021: 11th International Workshop on
  Bibliometric-enhanced Information Retrieval at ECIR 2021}, pages 20--36.

\bibitem[{Baker et~al.(2015)Baker, Silins, Guo, Ali, Högberg, Stenius, and
  Korhonen}]{10.1093/bioinformatics/btv585}
Simon Baker, Ilona Silins, Yufan Guo, Imran Ali, Johan Högberg, Ulla Stenius,
  and Anna Korhonen. 2015.
\newblock \href {https://doi.org/10.1093/bioinformatics/btv585} {{Automatic
  semantic classification of scientific literature according to the hallmarks
  of cancer}}.
\newblock \emph{Bioinformatics}, 32(3):432--440.

\bibitem[{Brambilla et~al.(2022)Brambilla, Rosi, Antici, Galassi, Giansanti,
  Magurano, Ruggeri, Torroni, Cisbani, and Lippi}]{10.3389/fpubh.2022.945181}
Gianfranco Brambilla, Antonella Rosi, Francesco Antici, Andrea Galassi, Daniele
  Giansanti, Fabio Magurano, Federico Ruggeri, Paolo Torroni, Evaristo Cisbani,
  and Marco Lippi. 2022.
\newblock \href {https://doi.org/10.3389/fpubh.2022.945181} {Argument mining as
  rapid screening tool of covid-19 literature quality: Preliminary evidence}.
\newblock \emph{Frontiers in Public Health}, 10.

\bibitem[{Butzke et~al.(2020)Butzke, Dulisch, Dunst, Steinfath, Neves, Mathiak,
  and Grune}]{Butzke2020SMAFIRAcAB}
Daniel Butzke, Nadine Dulisch, Sebastian Dunst, Matthias Steinfath, Mariana
  Neves, Brigitte Mathiak, and Barbara Grune. 2020.
\newblock \href {https://doi.org/10.21203/rs.3.rs-16454/v1} {Smafira-c: A
  benchmark text corpus for evaluation of approaches to relevance ranking and
  knowledge discovery in the biomedical domain}.
\newblock \emph{PREPRINT (Version 1) available at Research Square}.

\bibitem[{Chernodub et~al.(2019)Chernodub, Oliynyk, Heidenreich, Bondarenko,
  Hagen, Biemann, and Panchenko}]{chernodub-etal-2019-targer}
Artem Chernodub, Oleksiy Oliynyk, Philipp Heidenreich, Alexander Bondarenko,
  Matthias Hagen, Chris Biemann, and Alexander Panchenko. 2019.
\newblock \href {https://doi.org/10.18653/v1/P19-3031} {{TARGER}: Neural
  argument mining at your fingertips}.
\newblock In \emph{Proceedings of the 57th Annual Meeting of the Association
  for Computational Linguistics: System Demonstrations}, pages 195--200,
  Florence, Italy. Association for Computational Linguistics.

\bibitem[{Commission et~al.(2020{\natexlab{a}})Commission, Centre, Adcock,
  Novotny, Nic, Dibusz, Hynes, Marshall, and Gribaldo}]{doi/10.2760/725821}
European Commission, Joint~Research Centre, I~Adcock, T~Novotny, M~Nic,
  K~Dibusz, J~Hynes, L~Marshall, and L~Gribaldo. 2020{\natexlab{a}}.
\newblock \href {https://doi.org/doi/10.2760/725821} {\emph{Advanced non-animal
  models in biomedical research : respiratory tract diseases}}.
\newblock Publications Office of the European Union.

\bibitem[{Commission et~al.(2022{\natexlab{a}})Commission, Centre, Canals,
  Romania, P, Nic, Dibusz, Novotny, Busquet, Rossi, Straccia, Daskalopoulos,
  and Gribaldo}]{10.2760/7190}
European Commission, Joint~Research Centre, J~Canals, P~Romania, Belio-Mairal
  P, M~Nic, K~Dibusz, T~Novotny, F~Busquet, F~Rossi, M~Straccia,
  E~Daskalopoulos, and L~Gribaldo. 2022{\natexlab{a}}.
\newblock \href {https://doi.org/doi/10.2760/7190} {\emph{Advanced Non-animal
  Models in Biomedical Research - Immunogenicity testing for advanced therapy
  medicinal products}}.
\newblock Publications Office of the European Union.

\bibitem[{Commission et~al.(2022{\natexlab{b}})Commission, Centre, Capellini,
  Fanni, Gasparotti, Vignali, Celi, Cioffi, Positano, Haxhiademi, Costa,
  Landini, Daskalopoulos, Piergiovanni, and Gribaldo}]{doi/10.2760/94608}
European Commission, Joint~Research Centre, K~Capellini, B~Fanni, E~Gasparotti,
  E~Vignali, S~Celi, M~Cioffi, V~Positano, D~Haxhiademi, E~Costa, L~Landini,
  E~Daskalopoulos, M~Piergiovanni, and L~Gribaldo. 2022{\natexlab{b}}.
\newblock \href {https://doi.org/doi/10.2760/94608} {\emph{Advanced non-animal
  models in biomedical research : cardiovascular diseases}}.
\newblock Publications Office of the European Union.

\bibitem[{Commission et~al.(2022{\natexlab{c}})Commission, Centre, Otero,
  Canals, Belio-Mairal, Nic, Dibusz, Novotny, Busquet, Rossi, Gastaldello,
  Gribaldo, and Straccia}]{doi/10.2760/617688}
European Commission, Joint~Research Centre, M~Otero, J~Canals, P~Belio-Mairal,
  M~Nic, K~Dibusz, T~Novotny, F~Busquet, F~Rossi, A~Gastaldello, L~Gribaldo,
  and M~Straccia. 2022{\natexlab{c}}.
\newblock \href {https://doi.org/doi/10.2760/617688} {\emph{Advanced non-animal
  models in biomedical research : autoimmune diseases}}.
\newblock Publications Office of the European Union.

\bibitem[{Commission et~al.(2021{\natexlab{a}})Commission, Centre, Romania,
  Folgiero, Nic, Dibusz, Novotny, Busquet, Rossi, Straccia, Daskalopoulos, and
  Gribaldo}]{doi/10.2760/393670}
European Commission, Joint~Research Centre, P~Romania, V~Folgiero, M~Nic,
  K~Dibusz, T~Novotny, F~Busquet, F~Rossi, M~Straccia, E~Daskalopoulos, and
  L~Gribaldo. 2021{\natexlab{a}}.
\newblock \href {https://doi.org/doi/10.2760/393670} {\emph{Advanced non-animal
  models in biomedical research : immuno-oncology}}.
\newblock Publications Office of the European Union.

\bibitem[{Commission et~al.(2020{\natexlab{b}})Commission, Centre, Rossi,
  Caforio, Nic, Dibusz, Folgiero, Romania, Novotny, Busquet, Straccia, and
  Gribaldo}]{doi/10.2760/618741}
European Commission, Joint~Research Centre, F~Rossi, M~Caforio, M~Nic,
  K~Dibusz, V~Folgiero, P~Romania, T~Novotny, F~Busquet, M~Straccia, and
  L~Gribaldo. 2020{\natexlab{b}}.
\newblock \href {https://doi.org/doi/10.2760/618741} {\emph{Advanced non-animal
  models in biomedical research : breast cancer}}.
\newblock Publications Office of the European Union.

\bibitem[{Commission et~al.(2021{\natexlab{b}})Commission, Centre, Witters,
  Verstraelen, Aerts, Miccoli, Delahanty, and Gribaldo}]{doi/10.2760/386}
European Commission, Joint~Research Centre, H~Witters, S~Verstraelen, L~Aerts,
  B~Miccoli, A~Delahanty, and L~Gribaldo. 2021{\natexlab{b}}.
\newblock \href {https://doi.org/doi/10.2760/386} {\emph{Advanced non-animal
  models in biomedical research : neurodegenerative diseases}}.
\newblock Publications Office of the European Union.

\bibitem[{Fergadis et~al.(2021)Fergadis, Pappas, Karamolegkou, and
  Papageorgiou}]{fergadis-etal-2021-argumentation}
Aris Fergadis, Dimitris Pappas, Antonia Karamolegkou, and Haris Papageorgiou.
  2021.
\newblock \href {https://doi.org/10.18653/v1/2021.argmining-1.10}
  {Argumentation mining in scientific literature for sustainable development}.
\newblock In \emph{Proceedings of the 8th Workshop on Argument Mining}, pages
  100--111, Punta Cana, Dominican Republic. Association for Computational
  Linguistics.

\bibitem[{Gu et~al.(2021)Gu, Tinn, Cheng, Lucas, Usuyama, Liu, Naumann, Gao,
  and Poon}]{pubmedbert}
Yu~Gu, Robert Tinn, Hao Cheng, Michael Lucas, Naoto Usuyama, Xiaodong Liu,
  Tristan Naumann, Jianfeng Gao, and Hoifung Poon. 2021.
\newblock \href {https://doi.org/10.1145/3458754} {Domain-specific language
  model pretraining for biomedical natural language processing}.
\newblock \emph{{ACM} Transactions on Computing for Healthcare}, 3(1):1--23.

\bibitem[{Hobbs et~al.(2021)Hobbs, Goralski, Mitchell, Simpson, Leka, Kotey,
  Sekira, Munro, Nadendla, Jackson, Gonzalez-Aguirre, Krallinger, Giglio, and
  Erill}]{10.3389/frma.2021.674205}
Elizabeth~T. Hobbs, Stephen~M. Goralski, Ashley Mitchell, Andrew Simpson,
  Dorjan Leka, Emmanuel Kotey, Matt Sekira, James~B. Munro, Suvarna Nadendla,
  Rebecca Jackson, Aitor Gonzalez-Aguirre, Martin Krallinger, Michelle Giglio,
  and Ivan Erill. 2021.
\newblock \href {https://doi.org/10.3389/frma.2021.674205} {Eco-collectf: A
  corpus of annotated evidence-based assertions in biomedical manuscripts}.
\newblock \emph{Frontiers in Research Metrics and Analytics}, 6.

\bibitem[{Jin and Szolovits(2018)}]{jin-szolovits-2018-hierarchical}
Di~Jin and Peter Szolovits. 2018.
\newblock \href {https://doi.org/10.18653/v1/D18-1349} {Hierarchical neural
  networks for sequential sentence classification in medical scientific
  abstracts}.
\newblock In \emph{Proceedings of the 2018 Conference on Empirical Methods in
  Natural Language Processing}, pages 3100--3109, Brussels, Belgium.
  Association for Computational Linguistics.

\bibitem[{Katti and Katti(2017)}]{KATTI2017125}
Dinesh~R. Katti and Kalpana~S. Katti. 2017.
\newblock \href {https://doi.org/10.1016/j.jmbbm.2017.05.030} {Cancer cell
  mechanics with altered cytoskeletal behavior and substrate effects: A 3d
  finite element modeling study}.
\newblock \emph{Journal of the Mechanical Behavior of Biomedical Materials},
  76:125--134.
\newblock Structure-Property Relationships in Biological and Bioinspired
  Materials.

\bibitem[{Kosach et~al.(2018)Kosach, Shkarina, Kravchenko, Tereshchenko,
  Kovalchuk, Skoroda, Krotevych, and
  Khoruzhenko}]{10.12688/f1000research.15447.2}
V~Kosach, K~Shkarina, A~Kravchenko, Y~Tereshchenko, E~Kovalchuk, L~Skoroda,
  M~Krotevych, and A~Khoruzhenko. 2018.
\newblock \href {https://doi.org/10.12688/f1000research.15447.2}
  {Nucleocytoplasmic distribution of s6k1 depends on the density and motility
  of mcf-7 cells in vitro [version 2; peer review: 2 approved]}.
\newblock \emph{F1000Research}, 7(1332).

\bibitem[{Krallinger et~al.(2005)Krallinger, Erhardt, and
  Valencia}]{KRALLINGER2005439}
Martin Krallinger, Ramon Alonso-Allende Erhardt, and Alfonso Valencia. 2005.
\newblock \href {https://doi.org/10.1016/S1359-6446(05)03376-3} {Text-mining
  approaches in molecular biology and biomedicine}.
\newblock \emph{Drug Discovery Today}, 10(6):439--445.

\bibitem[{Larsson et~al.(2017)Larsson, Baker, Silins, Guo, Stenius, Korhonen,
  and Berglund}]{10.1371/journal.pone.0173132}
Kristin Larsson, Simon Baker, Ilona Silins, Yufan Guo, Ulla Stenius, Anna
  Korhonen, and Marika Berglund. 2017.
\newblock \href {https://doi.org/10.1371/journal.pone.0173132} {Text mining for
  improved exposure assessment}.
\newblock \emph{PLOS ONE}, 12(3):1--21.

\bibitem[{Lauscher et~al.(2018)Lauscher, Glava{\v{s}}, and
  Eckert}]{lauscher-etal-2018-arguminsci}
Anne Lauscher, Goran Glava{\v{s}}, and Kai Eckert. 2018.
\newblock \href {https://doi.org/10.18653/v1/W18-5203} {{A}rgumin{S}ci: A tool
  for analyzing argumentation and rhetorical aspects in scientific writing}.
\newblock In \emph{Proceedings of the 5th Workshop on Argument Mining}, pages
  22--28, Brussels, Belgium. Association for Computational Linguistics.

\bibitem[{Lippi et~al.(2022)Lippi, Antici, Brambilla, Cisbani, Galassi,
  Giansanti, Magurano, Rosi, Ruggeri, and Torroni}]{ijcai2022p857}
Marco Lippi, Francesco Antici, Gianfranco Brambilla, Evaristo Cisbani, Andrea
  Galassi, Daniele Giansanti, Fabio Magurano, Antonella Rosi, Federico Ruggeri,
  and Paolo Torroni. 2022.
\newblock \href {https://doi.org/10.24963/ijcai.2022/857} {Amica: An
  argumentative search engine for covid-19 literature}.
\newblock In \emph{Proceedings of the Thirty-First International Joint
  Conference on Artificial Intelligence, {IJCAI-22}}, pages 5932--5935.
  International Joint Conferences on Artificial Intelligence Organization.
\newblock Demo Track.

\bibitem[{Lippi and Torroni(2016)}]{LIPPI2016292}
Marco Lippi and Paolo Torroni. 2016.
\newblock \href {https://doi.org/10.1016/j.eswa.2016.08.050} {Margot: A web
  server for argumentation mining}.
\newblock \emph{Expert Systems with Applications}, 65:292--303.

\bibitem[{Luo et~al.(2022)Luo, Lai, Wei, Arighi, and Lu}]{10.1093/bib/bbac282}
Ling Luo, Po-Ting Lai, Chih-Hsuan Wei, Cecilia~N Arighi, and Zhiyong Lu. 2022.
\newblock \href {https://doi.org/10.1093/bib/bbac282} {{BioRED: a rich
  biomedical relation extraction dataset}}.
\newblock \emph{Briefings in Bioinformatics}, 23(5).
\newblock Bbac282.

\bibitem[{Mayer et~al.(2019)Mayer, Cabrio, and Villata}]{ijcai2019p953}
Tobias Mayer, Elena Cabrio, and Serena Villata. 2019.
\newblock \href {https://doi.org/10.24963/ijcai.2019/953} {Acta a tool for
  argumentative clinical trial analysis}.
\newblock In \emph{Proceedings of the Twenty-Eighth International Joint
  Conference on Artificial Intelligence, {IJCAI-19}}, pages 6551--6553.
  International Joint Conferences on Artificial Intelligence Organization.

\bibitem[{Menin et~al.(2021)Menin, Michel, Gandon, Gazzotti, Cabrio, Corby,
  Giboin, Marro, Mayer, Villata, and Winckler}]{10.1162/qss_a_00164}
Aline Menin, Franck Michel, Fabien Gandon, Raphaël Gazzotti, Elena Cabrio,
  Olivier Corby, Alain Giboin, Santiago Marro, Tobias Mayer, Serena Villata,
  and Marco Winckler. 2021.
\newblock \href {https://doi.org/10.1162/qss_a_00164} {{Covid-on-the-Web:
  Exploring the COVID-19 scientific literature through visualization of linked
  data from entity and argument mining}}.
\newblock \emph{Quantitative Science Studies}, 2(4):1301--1323.

\bibitem[{Pradhan et~al.(2018)Pradhan, Smith, Garson, Hassani, Seeto, Pant,
  Arnold, Prabhakarpandian, and Lipke}]{Pradhan2018}
Shantanu Pradhan, Ashley~M. Smith, Charles~J. Garson, Iman Hassani, Wen~J.
  Seeto, Kapil Pant, Robert~D. Arnold, Balabhaskar Prabhakarpandian, and
  Elizabeth~A. Lipke. 2018.
\newblock \href {https://doi.org/10.1038/s41598-018-21075-9} {A
  microvascularized tumor-mimetic platform for assessing anti-cancer drug
  efficacy}.
\newblock \emph{Scientific Reports}, 8(1):3171.

\bibitem[{Ruosch et~al.(2022)Ruosch, Sarasua, and Bernstein}]{zora217835}
Florian Ruosch, Cristina Sarasua, and Abraham Bernstein. 2022.
\newblock \href {https://doi.org/10.5167/uzh-217835} {Bam: Benchmarking
  argument mining on scientific documents}.
\newblock In \emph{The AAAI-22 Workshop on Scientific Document Understanding at
  the Thirty-Sixth AAAI Conference on Artificial Intelligence (AAAI-22)}. CEUR
  Workshop Proceedings.

\bibitem[{Stab et~al.(2018)Stab, Daxenberger, Stahlhut, Miller, Schiller,
  Tauchmann, Eger, and Gurevych}]{stab-etal-2018-argumentext}
Christian Stab, Johannes Daxenberger, Chris Stahlhut, Tristan Miller, Benjamin
  Schiller, Christopher Tauchmann, Steffen Eger, and Iryna Gurevych. 2018.
\newblock \href {https://doi.org/10.18653/v1/N18-5005} {{A}rgumen{T}ext:
  Searching for arguments in heterogeneous sources}.
\newblock In \emph{Proceedings of the 2018 Conference of the North {A}merican
  Chapter of the Association for Computational Linguistics: Demonstrations},
  pages 21--25, New Orleans, Louisiana. Association for Computational
  Linguistics.

\bibitem[{Trautmann et~al.(2020)Trautmann, Daxenberger, Stab, Schuetze, and
  Gurevych}]{epub88715}
Dietrich Trautmann, Johannes Daxenberger, Christian Stab, Hinrich Schuetze, and
  Iryna Gurevych. 2020.
\newblock \href {https://ojs.aaai.org/index.php/AAAI/article/view/6438}
  {Fine-grained argument unit recognition and classification}.
\newblock \emph{Thirty-Fourth Aaai Conference on Artificial Intelligence, the
  Thirty-Second Innovative Applications of Artificial Intelligence Conference
  and the Tenth Aaai Symposium on Educational Advances in Artificial
  Intelligence}, pages 9048--9056.

\bibitem[{Wachsmuth et~al.(2017)Wachsmuth, Potthast, Al-Khatib, Ajjour,
  Puschmann, Qu, Dorsch, Morari, Bevendorff, and
  Stein}]{wachsmuth-etal-2017-building}
Henning Wachsmuth, Martin Potthast, Khalid Al-Khatib, Yamen Ajjour, Jana
  Puschmann, Jiani Qu, Jonas Dorsch, Viorel Morari, Janek Bevendorff, and Benno
  Stein. 2017.
\newblock \href {https://doi.org/10.18653/v1/W17-5106} {Building an argument
  search engine for the web}.
\newblock In \emph{Proceedings of the 4th Workshop on Argument Mining}, pages
  49--59, Copenhagen, Denmark. Association for Computational Linguistics.

\bibitem[{Wang et~al.(2022)Wang, Song, Zhou, and Cheng}]{10.1002/asi.24590}
Xiaoguang Wang, Ningyuan Song, Huimin Zhou, and Hanghang Cheng. 2022.
\newblock \href {https://doi.org/10.1002/asi.24590} {The representation of
  argumentation in scientific papers: A comparative analysis of two research
  areas}.
\newblock \emph{J. Assoc. Inf. Sci. Technol.}, 73(6):863–878.

\bibitem[{Wei et~al.(2019)Wei, Allot, Leaman, and Lu}]{10.1093/nar/gkz389}
Chih-Hsuan Wei, Alexis Allot, Robert Leaman, and Zhiyong Lu. 2019.
\newblock \href {https://doi.org/10.1093/nar/gkz389} {{PubTator central:
  automated concept annotation for biomedical full text articles}}.
\newblock \emph{Nucleic Acids Research}, 47(W1):W587--W593.

\end{thebibliography}
